\documentclass[10pt,journal,compsoc]{IEEEtran}
\ifCLASSOPTIONcompsoc
  \usepackage[nocompress]{cite}
\else
  \usepackage{cite}
\fi

\usepackage{amsmath}
\usepackage{amsfonts,color}
\usepackage{mathrsfs}
\usepackage{latexsym,amssymb}
\usepackage{graphicx}
\usepackage{cite}
\usepackage{subfigure,balance}
\usepackage{fancyhdr}  
\usepackage{graphicx}
\usepackage{algorithmic}
\usepackage{graphicx}
\usepackage{textcomp}
\usepackage{hyperref}
\usepackage{stfloats}
\usepackage{float}
\usepackage{amsmath}
\usepackage{amsfonts,color}
\usepackage{mathrsfs}
\usepackage{latexsym,amssymb}
\usepackage{graphicx}
\usepackage{cite}
\usepackage{subfigure,balance}
\usepackage{fancyhdr}
\usepackage{makecell}
\usepackage{booktabs}
\usepackage{algorithmic}
\usepackage{booktabs}
\usepackage{ragged2e}
\usepackage{tabularx}
\usepackage{bbding}
\usepackage{pifont}
\usepackage{wasysym}
\usepackage{amssymb}

\usepackage{enumitem,amssymb}
\newlist{todolist}{itemize}{2}
\setlist[todolist]{label=$\square$}
\usepackage{pifont}
\ifCLASSINFOpdf
\else
\fi
\begin{document}
%
\title{A Lite Fireworks Algorithm  for Optimization}
%
%
%
%

\author{Haimiao~Mo*, Min~Zeng  
\IEEEcompsocitemizethanks{
\IEEEcompsocthanksitem 

\textbf{Haimiao Mo*} is with the School of Management, Hefei University of Technology, Anhui Hefei 23009, China, also with the Key Laboratory of Process Optimization and Intelligent Decision-Making, Ministry of Education, China.(Email: mhm\_hfut@163.com)
\protect\\

\IEEEcompsocthanksitem 
\textbf{Min Zeng} is with College of Computer Science, Guangdong University of Science and Technology, Dongguan, Guangdong Province, China.(Email: zm6102@163.com)
 } }

\IEEEtitleabstractindextext{%
\begin{abstract}
\justifying 
The fireworks algorithm is an optimization algorithm for simulating the explosion phenomenon of fireworks. Because of its fast convergence and high precision, it is widely used in pattern recognition, optimal scheduling, and other fields. However, most of the existing research work on the fireworks algorithm is improved based on its defects, and little consideration is given to reducing the number of parameters of the fireworks algorithm. The original fireworks algorithm has too many parameters, which increases the cost of algorithm adjustment and is not conducive to engineering applications. In addition, in the fireworks population, the unselected individuals are discarded, thus causing a waste of their location information. To reduce the number of parameters of the original Fireworks Algorithm and make full use of the location information of discarded individuals, we propose a simplified version of the Fireworks Algorithm. It reduces the number of algorithm parameters by redesigning the explosion operator of the fireworks algorithm and constructs an adaptive explosion radius by using the historical optimal information to balance the local mining and global exploration capabilities. The comparative experimental results of function optimization show that the overall performance of our proposed LFWA is better than that of comparative algorithms, such as the fireworks algorithm, particle swarm algorithm, and bat algorithm.

\end{abstract}

}

\maketitle

\IEEEdisplaynontitleabstractindextext

%
\IEEEpeerreviewmaketitle

\IEEEraisesectionheading{\section{Introduction}\label{sec:introduction}}

%
%
%
%

\IEEEPARstart{I}{nspired} by the natural phenomenon that fireworks explode in the night sky to generate sparks and illuminate the surrounding area, Tan et al. proposed the Fireworks Algorithm (FWA) in 2010 \cite{tan2010fireworks}. In this algorithm, fireworks are regarded as a feasible solution in the solution space of the optimization problem, so the process of fireworks explosion to produce a certain number of sparks is the process of searching the neighborhood. The fireworks algorithm is widely used in pattern recognition \cite{zhang2020recognition}, optimal scheduling \cite{he2019discrete}, function optimization \cite{cheng2019improved}, and other domains because it has fast convergence and high precision and can balance local mining and global exploration capabilities through exploding sparks.

However, according to previous studies, there are some flaws in the design of the original fireworks algorithm  \cite{li2019comprehensive}. First, when the fireworks of the original fireworks algorithm explode, each dimension produces the same displacement. It may lead to insufficient diversity of the population, which in turn limits the exploration ability of the algorithm.
Second, the original fireworks select the next-generation fireworks through a strategy based on distance, which leads to inefficiency in selecting offspring fireworks and increased computational cost. Third, the original firework algorithm has different subpopulations, such as fireworks subpopulation, explosion spark subpopulation, explosion spark subpopulation, and Gaussian spark subpopulation. The local mining and global exploration capacities of the fireworks algorithm are constrained by the absence of information-sharing and coordination mechanisms among these subpopulations.

Improvement efforts for this algorithm are mostly classified into two categories based on the shortcomings of the original fireworks method \cite{zhigang2022review}. The first type of improvement work is to improve explosion operators (such as explosion strength, explosion radius, and displacement operations), mutation operators, mapping rules, and selection strategies. Another type of improvement work is to improve the fireworks algorithm by combining the advantages of other heuristic algorithms \cite{li2019comprehensive}. 

In the first type of improvement work, there are three main improvement strategies. First, the explosion operator (mutation operator) is improved by using the effective information of fireworks (mapping rules, or pros and cons of explosion spark difference vector) to enhance the local mining and global exploration capabilities (increase population diversity) of the fireworks algorithm. This improvement strategy mainly enhances the optimization ability of the algorithm by improving the explosion radius of the fireworks algorithm \cite{li2014adaptive}, \cite{zheng2014dynamic}, \cite{zheng2015cooperative}, \cite{pekdemir2016enhancing} random mapping rules or boundary mapping rules \cite{cheng2015analytics}, \cite{ye2018mapping}, and population diversity \cite{li2015orienting}. Second, due to the low efficiency and high cost of selecting offspring, this type of improvement strategy is mainly improved for the offspring selection strategy of the fireworks algorithm \cite{pei2012empirical}, \cite{jun2019fireworks}, \cite{chen2021explosion}, such as "random-elite" strategy, and tournament strategy. Third, the improved firework algorithm establishes an effective collaborative optimization mechanism between sub-populations through information exchange between sub-populations \cite{zheng2015cooperative}, \cite{lv2022novel}, \cite{liu2021neighborhood}, \cite{li2016enhancing}, \cite{li2006tan}. It can not only improve the information utilization efficiency among the subpopulations of the original fireworks algorithm, and select excellent subpopulations to guide the population to find the optimal solution, but also effectively enhance the optimization ability of the algorithm.

The second type of improvement work mainly enhances the performance of the hybrid algorithm by introducing the advantages of other evolutionary algorithms, such as Particle Swarm Optimization (PSO) \cite{zhu2021novel}, Moth Search Algorithm (MSA)\cite{han2020efficient}, Gray Wolf Optimizer (GWO) \cite{yue2020novel}, Biogeography-based Optimization Algorithm (BOA)\cite{zhang2014hybrid}, Genetic Algorithm (GA) \cite{lapa2016application}, \cite{babu2016parameter}. The original fireworks algorithm's local mining and global exploration capabilities are significantly enhanced by integrating the benefits of these evolutionary algorithms.

However, most of the existing research work is to improve the defects of the fireworks algorithm and rarely considers reducing the number of parameters of the fireworks algorithm.  When calculating the explosion intensity, the original firework algorithm \cite{tan2010fireworks} needs to use a constant to control the number of explosions sparks. The original fireworks algorithm requires predefining the threshold of the maximum explosion radius before computing the explosion radius. The original fireworks algorithm has too many parameters, which increases the cost of parameter adjustment and is not conducive to the wide application of the algorithm.

In addition, when the fireworks algorithm selects offspring fireworks, individuals that are not selected as the next generation will be discarded, which greatly wastes the historical information of these individuals. We simplify the parameters of the fireworks algorithm and make full use of the historical information of the unselected individuals to construct an adaptive explosion radius, to balance the local mining and global exploration capabilities of the algorithm. 

Therefore, we propose a Lite Fireworks Algorithm (LFWA) to reduce the number of parameters of the original fireworks algorithm and adaptively adjust the explosion radius. The contributions of our LFWA are as follows.

1) By redesigning the explosion operator of the fireworks algorithm, the algorithm is simplified to reduce the number of parameters of the algorithm, which is more conducive to the parameter adjustment and application practice of the algorithm.

2) By using the historical optimal information of the fireworks algorithm to construct a new explosion radius, the fireworks population can better balance the local search and global search capabilities to improve the performance of the algorithm.

\section{The Lite Fireworks Algorithm}

\subsection{Explosion Intensity}

Since the original fireworks algorithm needs to predefine the maximum number of explosion sparks, too many parameters may harm the application of the algorithm. Equation (1) illustrates the new strategy for calculating explosion intensity, which reduces the number of parameters of the fireworks algorithm. It limits the maximum number of explosion sparks by the fitness value of individual fireworks and the population size of fireworks.

\begin{equation}
	{S_i} = \left\lceil {{M^{\frac{{{f_{\max }} - f({x_i})}}{{{f_{\max }} - {f_{\min }} + \xi }}}}} \right\rceil
\end{equation}
where $S_{i}$ is the explosion intensity of the fireworks $x_i$, $M$ is the population of fireworks. $\left\lceil {} \right\rceil $ is a ceil function to limit the number of $S_{i}$ in the integer range of $[1, M]$.

\subsection{Explosion Radius}

When selecting the next generation of fireworks, the original fireworks algorithm will discard the unselected individuals, which greatly wastes the historical information of these discarded individuals. Therefore, the historical information of these individual locations is used to construct a new explosion radius to balance the local mining and global exploration capabilities of the fireworks algorithm. The new explosion radius is calculated as shown in Equations (2) and (3).

\begin{equation}
	{R_i} = \left\{ \begin{array}{l}
		pbes{t_i} - {x_i}, {S_i} < {S_{Avg}}\\
		x_{CF} - x{_i}, {S_i} \ge {S_{Avg}}
	\end{array} \right.
\end{equation}
\begin{equation}
{S_{Avg}} = \frac{1}{M}\sum\limits_{i = 1}^M {{S_i}} 
\end{equation}
where $pbest_{i}$ is the historical best location information of the \textit{i}-th fireworks $x_{i}$, and $x_{CF}$ is the best fireworks among all $pbest = \{ pbes{t_1},pbes{t_2},...,pesbt{}_M\}$, also called Core Fireworks (CF). $S_{Avg}$ is the average explosion intensity, which is also the average number of all explosion sparks.

\subsection{Displacement Operation}

When exploding, the fireworks $x_i$ complete the displacement and then generates explosion sparks with the number of $S_i$. The displacement operation is given by Equation (4).

\begin{equation}
	E{S_j} = {x_i} + \beta  \times {R_i}
\end{equation}
where $j=1,2, ..., S{_i}$. And $x_i$ is the position of \textit{i}-th fireworks, and $ES_j$ is the \textit{j}-th Explosion Sparks (ES) in the \textit{i}-th subpopulation generated by the \textit{i}-th fireworks $x_i$ after an explosion. $\beta$ is a random number with a uniform distribution in the range [0,1], $R_i$ is the explosion radius of the \textit{i}-th fireworks $x_i$.

From Equations (2) to (4), LFWA adopts $x_{CF}$ and $pbest$ to construct a new explosion radius that can adaptively adjust the explosion radius according to the explosion intensity. If the explosion intensity of the \textit{i}-th fireworks $x_i$ is less than the average explosion intensity $S_{Avg}$, then the fireworks learn from $pbest$. Otherwise, the fireworks learn from the core fireworks $x_{CF}$. In this way, LFWA can adaptively adjust the step size by introducing $x_{CF}$ and $pbest$ to construct a new explosion radius.

\subsection{Mutation Factor}

The LFWA maintains the diversity of the fireworks population through the mutation strategy, so as to avoid the fireworks population falling into the local optimal solution too quickly. The mutation strategy of LFWA is shown in Equation (5), which randomly selects \textit{n} dimensions of fireworks for Gaussian mutation. Gaussian Sparks (GS) is generated after the fireworks undergo Gaussian mutation.

\begin{equation}
	G{S_{ij}} = {x_{ij}} \times (N(0,1) + 1)
\end{equation}
where $j =1,2,..., n$. $N(0, 1)$ is a Gaussian distribution function with a mean of 0 and a standard deviation of 1. $d$ is the total dimensions of $x_i$.

\subsection{Mapping Rules}

Each dimension of the positions of fireworks, explosion sparks, and Gaussian sparks needs to be processed using the mapping rules of Equation (6) to prevent them from going out of bounds.

\begin{equation}
{x_{ij}^{new}} = LB + \beta  \times (UB - LB)
\end{equation}
where $UB$ and $LB$ are the upper and lower bounds of the algorithm's search range, respectively.

\subsection{Selection Strategy}

The LFWA adopts the "Elite-Random" strategy \cite{pei2012empirical} to select the next generation of fireworks from the candidate set, which consists of fireworks $x = \{ x{}_1,{x_2},...,x{}_M\}$, $pbest$, $x_{CF}$, explosion sparks $ES$, and Gaussian fireworks $GF$. That is to say, the core firework with the best fitness is selected as the next generation of fireworks individuals, and the other fireworks individuals of the next generation are randomly selected.

\section{Comparative Experiments}

\subsection{Benchmark Functions}

Benchmark functions are utilized in comparison tests to examine how well the suggested LFWA performs. Table 1 displays the function expression.

\begin{table*}[!htbp]
	\centering
	\caption{Benchmark functions}
	\label{TABLE1}
	\begin{tabularx}{\textwidth}{p{2.9cm}p{9.2cm}p{2.4cm}p{0.5cm}p{1.8cm}}
		\toprule
		Function Name &  Function & Domain & dim & optimum \\
		\midrule
		Sphere & $ {f_1}(x) = \sum\limits_{i = 1}^n {{x_i}} $ & ${[ - 100, 100]^n}$ & 30 & 0 \\
		Rosenbrock & ${f_2}(x) = \sum\limits_{i = 1}^n {[100{{({x_{i + 1}} - {x_i})}^2} + {{({x_i} - 1)}^2}]}$ & ${[ - 10, 10]^n}$  & 30 & 0
		\\
		
		Rosenbrock & ${f_3}(x) = \sum\limits_{i = 1}^n {[x_i^2 - 10\cos (2\pi {x_i}) + 10]}$ & ${[ - 5.12, 5.12]^n}$  & 30 & 0
		\\
		Griewank & ${f_4}(x) = \frac{1}{{4000}}\sum\limits_{i = 1}^n {x_i^2 - \prod\limits_{i = 1}^n {\cos (\frac{{{x_i}}}{{\sqrt i }}) + 1} }$  &  ${[ - 600, 600]^n}$  & 30 &0
		\\
		
		Ackley & ${f_5}(x) =  - 20\exp ( - 0.2\sqrt {\sum\limits_{i = 1}^n {x_i^2} } ) - \exp (\frac{1}{n})\sum\limits_{i = 1}^n {\cos (2\pi {x_i}) + 20 + e} $  &  ${[ - 32, 32]^n}$  & 30 &0
		\\
		Schwefel & ${f_6}(x) = \sum\limits_{i = 1}^n { - {x_i}\sin (\sqrt {\left| {{x_i}} \right|} )}$ &  ${[ - 100, 100]^n}$  & 30 & 0
		\\
		Six-Hump Camel-Back & ${f_7}(x) = 4x_1^2 - 2.1x_1^4 + 3.1x_1^6 + {x_1}{x_2} - 4x_2^2 + 4x_2^4$ &  ${[ - 5, 5]^n}$  & 2 & -1.0316285
		\\
		Goldstein Price & ${f_8}(x) = [1 + {({x_1} + {x_2} + 1)^2}(19 - 14{x_1} + 3x_1^2 - 14{x_2} + 6{x_1}{x_2} + 3x_2^4)] \times [30{\rm{ + }}{(2{x_1} - 3x{}_2)^2}(18 - 32{x_1} + 12x_1^2 + 48{x_2} - 36{x_1}{x_2} + 27x_2^2)]$ &  ${[ - 2, 2]^n}$  & 2 & 3
		\\
		Schaffer’s F6 & ${f_9}(x)\frac{{{{\sin }^2}\sqrt {x_1^2 + x_2^2}  - 0.5}}{{{{[1 + 0.001(x_1^2 + x_2^2)]}^2}}} + 0.5$ &   ${[ - 100, 100]^n}$   & 2 & 0
		
		\\\midrule
	\end{tabularx}%
\end{table*}%

\subsection{Parameters Setting}

The performance of LFWA is compared with other algorithms (e.g. BA, SPSO, FWA) by testing the benchmark functions shown in Table 1. A fixed tolerance $\lambda  \le {10^{{\rm{ - }}5}}$ is used in experiments to measure the error between the search value and the optimal value of the objective function. If the error is less than or equal to $\lambda$, it means that the algorithm successfully finds the optimal solution.

\subsection{Analysis of Convergence}

\begin{table}[]
	\centering
	\caption{The comparative experimental results of the four algorithms}
	\begin{tabular}{@{}lllllllllll@{}}
		\toprule
		$f$  & Alg  & worst     & best      & mean       & SD          \\ \midrule
		$f_1$   &	LFWA   & \textbf{0}         & \textbf{0}         & \textbf{0}         & \textbf{0} \\
		&  BA & 1.698E-03 & 7.972E-04 & 1.367E-03 & 1.547E-04  \\
		&   SPSO & 7.170E+01  & 5.100E+00  &2.207E+01  & 1.164E+01  \\
		&  FWA  &1.625E-173& 4.067E-260 &1.625E-175 & \textbf{0}\\\midrule
		$f_2$   &	LFWA   & \textbf{2.883E+01} & 2.553E+01 & \textbf{2.749E+01} & 1.018E+00 \\ 
		&  BA & 8.094E+01 & \textbf{2.419E+01} & 2.810E+01 & 5.502E+00 \\
		&SPSO & 2.563E+05 & 2.344E+03 &  4.160E+04 & 4.982E+04\\ 
		& FWA & 2.890E+01 & 2.706E+01 & 2.852E+01 & \textbf{4.811E-01}\\ \midrule
		
		$f_3$   &	LFWA   & \textbf{0}         & \textbf{0}         & \textbf{0}         & \textbf{0} \\
		& BA & 9.397E-05 &  4.027E-05 &  7.027E-05 & 1.063E-05\\
		& SPSO & 8.792E-01 &  2.158E-01 &  5.811E-01 &  1.355E-01 \\
		&	FWA   & \textbf{0 }        & \textbf{0 }        & \textbf{0 }        & \textbf{0 }\\ \midrule
		
		$f_4$  &	LFWA   & \textbf{0 }        & \textbf{0  }       &\textbf{ 0 }        & \textbf{0} \\
		& BA  & 5.496E+01 & 1.410E+01 & 2.879E+01 & 8.334E+00 \\
		& SPSO & 6.407E+02 & 2.054E+02 & 3.339E+02 & 8.129E+01\\
		&	FWA   & \textbf{0  }       & \textbf{0 }        &\textbf{ 0   }      & \textbf{0} \\\midrule
		
		$f_5$  &	LFWA   &8.882E-16 & 8.882E-16  &8.882E-16 & 0\\
		& BA &3.225E+00  & 1.902E+00  & 2.733E+00  & 2.978E-01\\
		& SPSO  &2.048E+01 & 5.323E+00 & 1.903E+01 & 3.471E+00 \\
		& FWA & 8.882E-16  & 8.882E-16  & 8.882E-16  & 0 \\ \midrule
		$f_6$ & LFWA & 1.197E-08 &  6.414E-16 &  6.693E-10 &  1.448E-09 \\
		& BA & 1.197E-08 & 6.414E-16 & 6.693E-10  &1.448E-09 \\
		& SPSO & \textbf{0}  &  \textbf{0 } &\textbf{ 0 }  & \textbf{0} \\
		& FWA & 1.916E-10  & 6.322E-37  & 2.513E-12  & 1.974E-11 \\  \midrule
		$f_7$ & LFWA & \textbf{-1.0316285}  & \textbf{-1.0316285 } & \textbf{-1.0316285}  & 1.508E-12\\
		& BA & -1.0316282   &\textbf{-1.0316285}  & -1.0316284  & 6.612E-08\\
		& SPSO & \textbf{-1.0316285 }& \textbf{-1.0316285}  & \textbf{-1.0316285} & \textbf{1.558E-15} \\
		
		& FWA & -1.0316089  & \textbf{-1.0316285}  & -1.0316273 & 2.492E-06\\ \midrule
		$f_8$ & LFWA & \textbf{3.000000}  & \textbf{3.000000} & \textbf{3.000000 } & 7.177E-11\\
		& BA  &  84.000027  &  \textbf{3.000000}  &  10.560005  &  1.539E+01\\
		& SPSO & \textbf{3.000000} &  \textbf{3.000000 }&  \textbf{3.000000 }  &  \textbf{1.085E-15}\\
		&  FWA &  3.000236  &   \textbf{3.000000}  &  \textbf{3.000010} &   2.824E-05 \\ \midrule
		$f_9$ & LFWA & \textbf{0}  & \textbf{0}   & \textbf{0}  & \textbf{0}  \\
		& BA & 1.2699E-01 & 1.4234E-10 & 6.8488E-03 & 1.788E-02\\
		& SPSO & 9.7159E-03  & \textbf{0} &
		1.7489E-03  & 3.752E-03\\
		& FWA & \textbf{0}  & \textbf{0}   & \textbf{0}  & \textbf{0}  \\  
		
		\bottomrule
	\end{tabular}
\end{table}

The experiment results of worst, best, mean, and Standard Deviation (SD), with 1000 iterations and 100 runs, are shown as Table 2. From Table 2, the mean of LFWA is better than the other three algorithms for $f_1$ and $f_2$ at the end of the iteration, so the optimization accuracy of LFWA is better than the other three algorithms. For $f_3$, $f_4$, $f_5$ and $f_9$, the mean of LFWA is the same as FWA, but better than the other two algorithms, so the optimization accuracy of LFWA and FWA is better than the other two algorithms at the end of the iteration. For $f_6$, $f_7$, and $f_8$, the mean of LFWA is the same as SPSO, but better than the other two algorithms, so the optimization accuracy of LFWA and SPSO is better than the other two algorithms. In a word, the overall optimization accuracy of LFWA is better than the other three algorithms at the end of the iteration.

We only give partial evolution curves for four algorithms shown in Figure 1(a) to Figure 1(g) that the convergence speed of LFWA is better than the other three algorithms. The evolution curves except Figure 1(b) and Figure 1(g) are partly overlapping, and their ordinates are $log_{10}(Fitness)$. From the Figure 1(a) to Figure 1(g) except Figure 1(f), it can be seen that during the early phase of optimizing function, the evolution of LFWA is steeper than these of the other three algorithms, and the convergence speed of LFWA is better than other three algorithms. Meanwhile, the optimization accuracy of LFWA is better than the other three algorithms when the iteration is the same.

In Figure 1(f), it can be seen that the convergence speed of LFWA is better than that of BA but worse than the other two algorithms during the early phase of optimization. However, from the mean of $f_8$ shown in Table 2, it can be seen that at the end of the iteration, the optimization accuracy of LFWA is the same as SPSO and better than the other two algorithms. In summary, LFWA outperforms the other three algorithms in terms of overall optimization accuracy and convergence speed.

\subsection{Analysis of Robustness}

\begin{figure}[htbp]
	\includegraphics[width = 9 cm]{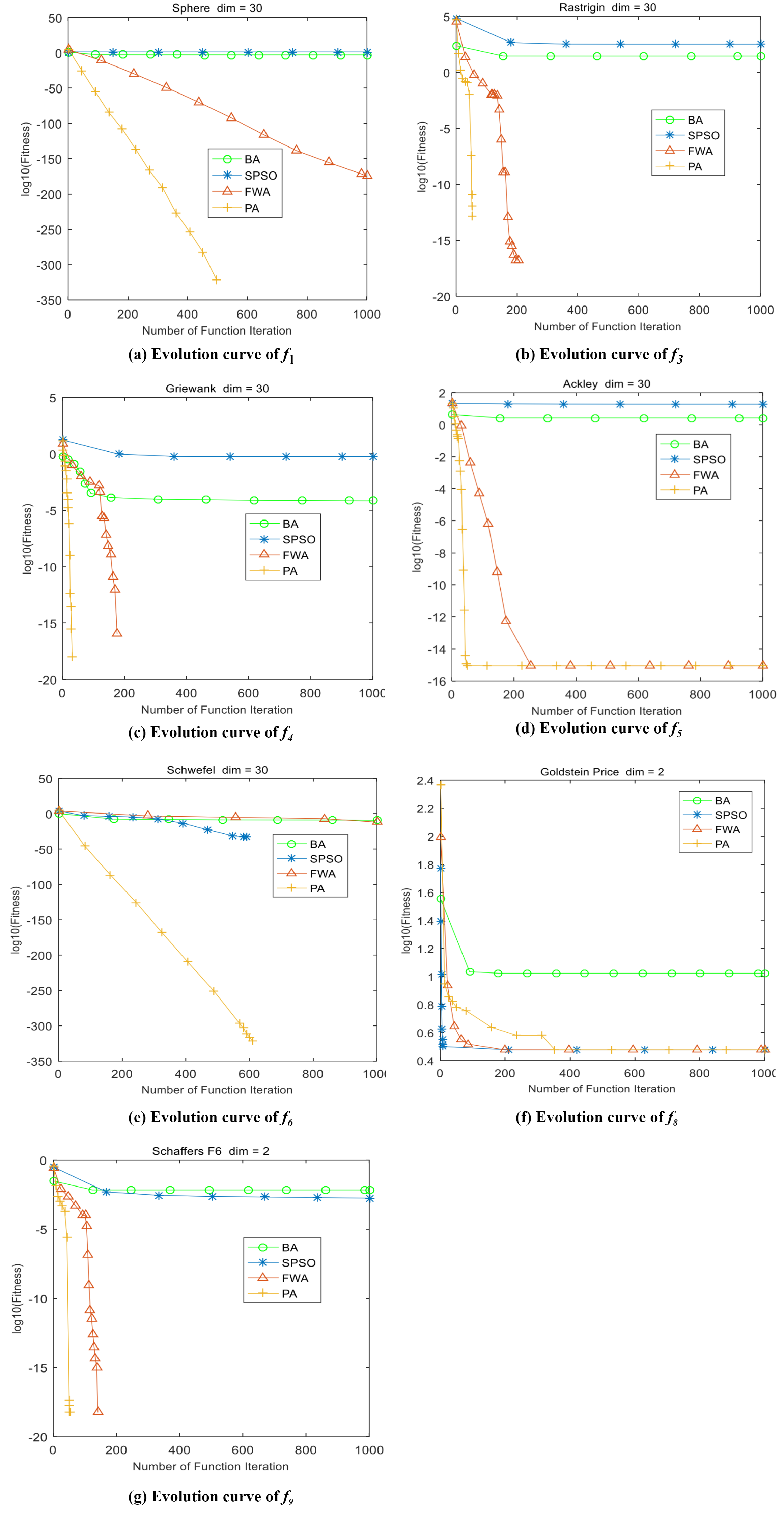}
	\caption{Evolution curves for partial functions of Table 1.}
	\label{Fig1}
\end{figure}

The standard deviation reflects the robustness of an algorithm. In other words, a smaller standard deviation presents better robustness for an algorithm, and reversely, presents worse robustness for an algorithm. From the standard deviation of Table 2, the robustness of LFWA is the same as FWA for $f_1$, $f_3$, $f_4$, $f_5$, and $f_9$, but better than the other two algorithms. For $f_2$, the robustness of LFWA is worse than FWA but better than the other two algorithms. For $f_6$, the robustness of LFWA is the same as SPSO, but better than the other two algorithms. For $f_2$, the robustness of LFWA is worse than FWA but better than the other two algorithms. For $f_6$, the robustness of LFWA is the same as SPSO, but better than the other two algorithms. In a word, the overall robustness of LFWA is better than the other three algorithms.

\subsection{Analysis of success rate}
\begin{table}[]
	\centering
	\caption{The success rate (\%) of four algorithms}
	\begin{tabular}{@{}lllllllllll@{}}
		\toprule
		alg &
		$f_1$ &
		$f_3$ &
		$f_4$ &
		$f_5$ &
		$f_6$ &
		$f_8$ &
		$f_9$ &
		$f_2$, 
		$f_7$    \\ \midrule
		LFWA  & \textbf{100}  & \textbf{100} & \textbf{100}  & \textbf{100}  & \textbf{100} & \textbf{100} & \textbf{100} &
		0  \\
		BA &
		0 &
		0 &
		0 &
		0 & \textbf{100} & 73  &67 &
		0  \\
		SPSO &
		0 &
		0 &
		0 &
		0 & \textbf{100}  & \textbf{100} & 82 &
		0  \\
		FWA  & \textbf{100} & \textbf{100}  &\textbf{100}  &\textbf{100} & \textbf{100} & 81 & \textbf{100} &
		0  \\

		\bottomrule
	\end{tabular}
\end{table}

If the accuracy satisfies the fixed tolerance $\lambda $ , it is recorded successfully once. The Success Rate (SR) for four algorithms are shown as Table 3. From Table 3, for $f_2$, $f_7$,  the success rate of the four algorithms is 0\%. For $f_1$ , $f_3$, $f_4$ and $f_5$, the success rate of LFWA and FWA is 100\%, they are better than other two algorithms. For $f_6$, the SR of the four algorithms is 100\%. For $f_8$, the SR of LFWA and SPSO is 100\%, and they are better than other two algorithms. For $f_9$, the SR of LFWA and FWA is 100\%, and they are better than other two algorithms. Therefore, the overall success rate of LFWA is better than other three algorithms for optimizing $f_1$ to $f_9$.

\subsection{Discussion}

As shown in the experiments, the LFWA has a faster convergence speed, a better optimization accuracy, a better robustness and a higher optimization success rate, compared to the BA, SPSO and FWA. The reason why LFWA outperforms other comparative algorithms lies in the following three aspects.

First, LFWA reduces the number of parameters of the original fireworks algorithm, and presents a more concise expression, which reduces the cost of parameter adjustment during the optimization process of the algorithm.

Second, LFWA draws on the memory mechanism of the particle swarm algorithm, and introduces the locally optimal fireworks individual $pbest$ and core fireworks $x_{CF}$ into the original fireworks algorithm during the optimization process. Then an adaptive explosion radius is constructed by utilizing the historical optimal information. In this way, LFWA can share and inherit the historical optimal information during the optimization process, so LFWA has a faster convergence speed.

Third, the "Elite-Rand" strategy and variation factors maintain the diversity of LFWA. Therefore, LFWA has a strong ability to avoid premature convergence.

\section{CONCLUSIONS}

Compared with the bat algorithm, standard particle swarm algorithm, and fireworks algorithm, we can conclude that the LFWA has a promising performance. The LFWA outperforms the other three algorithms on benchmark functions in terms of optimization accuracy, convergence speed, robustness, and optimization success rate, which endues the LFWA with a promising prospect of application and extension. In future work, we will seek a deep theoretical analysis of the LFWA and try to apply the LFWA to some special engineering applications.

\ifCLASSOPTIONcaptionsoff
  \newpage
\fi



%

\bibliographystyle{IEEEtran} 	
\bibliography{reference_LFWA_optimization}

\end{document}